\documentclass[sn-mathphys-num]{sn-jnl}

\usepackage{graphicx}%
\usepackage{multirow}%
\usepackage{amsmath,amssymb,amsfonts}%
\usepackage{amsthm}%
\usepackage{mathrsfs}%
\usepackage[title]{appendix}%
\usepackage{xcolor}%
\usepackage{textcomp}%
\usepackage{manyfoot}%
\usepackage{booktabs}%
\usepackage{algorithm}%
\usepackage{algorithmicx}%
\usepackage{algpseudocode}%
\usepackage{listings}%
\usepackage[dvipsnames]{xcolor}
\usepackage{soul}
\usepackage{comment}
\usepackage{bm}

\begin{document}

\title[Article Title]{A Sliding Mode Controller Based on Timoshenko Beam Theory
Developed for a Tendon-Driven Robotic Wrist }

\author*[1]{\fnm{Shifa} \sur{Sulaiman}}\email{ssajmech@gmail.com}
\author[1]{\fnm{Mohammad} \sur{Gohari}}
\author[1]{\fnm{Francesco} \sur{Schetter}}

\author[1]{\fnm{Fanny} \sur{Ficuciello}}

\affil[1]{\orgdiv{Department of Information Technology and Electrical Engineering}, \orgname{University of Naples Federico II}, \orgaddress{\street{Claudia}, \city{Naples}, \postcode{80125}, \state{Campania}, \country{Italy}}}

\abstract{ Development of dexterous robotic joints is essential for advancing manipulation capabilities in robotic systems. This paper presents a design and an implementation of a tendon-driven robotic wrist joint, together with an efficient Sliding Mode Controller (SMC) for precise motion control. The wrist mechanism is modelled using the Timoshenko-based approach to accurately capture its kinematic and dynamic properties, which serve as the foundation for tendon force calculations within the controller. The proposed SMC is designed to deliver fast dynamic response and computational efficiency, enabling accurate trajectory tracking under varying operating conditions.
%from here old abstract
%The advancement of prosthetic hand technology relies heavily on the development of dexterous robotic joints capable of replicating intricate human wrist movements. This paper focuses on the design and implementation of a dexterous tendon-driven robotic wrist joint for prosthetic applications, coupled with the development of an efficient sliding mode controller (SMC) for manipulating the motions of the robotic wrist joint.The proposed robotic wrist joint described in this work lays the groundwork for future wrist designs that enhance dexterity, adaptability,  and precision, enabling seamless integration with the human wrist and bridging the gap between artificial and natural movements.
 The effectiveness of the proposed controller is validated through comparative analyses with existing controllers for similar wrist mechanisms. The proposed SMC demonstrated superior performance, validated through both simulation and experimental studies. The Root Mean Square Error (RMSE) range in simulation is found to be approximately 
$1.67 \times 10^{-2}$ radians, while experimental validation yielded an error of $0.2$ radians. Additionally, the controller achieved a settling time of less than $3$ seconds and a steady-state error below 
$ 10^{-1} $ radians, consistently observed across both simulation and experimental evaluations. Comparative analyses with other controllers confirmed that the developed SMC surpassed alternative control strategies in motion accuracy, rapid convergence, and steady-state precision, contributing to enhanced dexterity. This work establishes a foundation for future exploration of tendon-driven wrist mechanisms and control strategies in robotic applications. %While the current study focuses on mechanism design and control validation, subsequent research may investigate compact actuation and user-centered evaluation to assess potential applicability in wearable or assistive contexts.

}

\keywords{Tendon-Driven robot, Robotic wrist, Sliding mode controller, Timoshenko beam theory}

\maketitle

\section{Introduction }

The incorporation of tendons allows soft robots to perform intricate movements while preserving their structural integrity \cite{soft,soft1}. Tendon-driven soft continuum robots present several significant benefits that improve the overall functionality and adaptability of the system \cite{wockenfuss2022design, tutcu2021quasi}. A range of modeling approaches, including Constant Curvature (CC) models \cite{escande}, Variable Curvature (VC) models \cite{huang}, Finite Element Method (FEM) \cite{xavier}, and Machine Learning (ML) techniques \cite{renda,ML1,ML2} are utilized to develop both static and dynamic models of soft robotic tendon driven systems. In this work, we utilised Timoshenko beam theory to model the tendon-driven robotic wrist. 

Analytical formulations of the kinematics and dynamics models of a continuum soft robotic arm utilizing Timoshenko beam theory were given in  Ref. \cite{tim1}. The findings indicated that the Timoshenko beam theory provided a more comprehensive and precise approach compared to alternative beam models.
A continuum robot inspired by centipedes and polychaete worms was presented in Ref. \cite{tim2}. This model emerged from the limit of a rigid body chain with interconnected components, resulting in a Timoshenko beam model characterized by a one-dimensional continuum featuring a local Euclidean structure. This structure effectively represented the cross sections, which were kinematically defined by their positions and orientations. A study exploring the kinetostatic and dynamic formulation of planar-compliant mechanisms through the application of the dynamic stiffness method grounded in Timoshenko beam theory was proposed in  Ref. \cite{tim3}. This investigation was motivated by the importance of accounting for shear deformation and rotary inertia in short and thick flexure beams commonly utilized in compliant mechanisms. The research involved the creation of a frequency-dependent dynamic stiffness matrix that incorporated pseudo-static characteristics to achieve three specific objectives.

A versatile bio-inspired slender mechanism, represented as a Timoshenko beam was presented in  Ref. \cite{tim4}. This mechanism interacted with its environment through a continuous array of compliant elements. A reduced order model was developed by projecting the governing partial differential equations onto the linear modal basis of the Timoshenko beam. The interaction with the substrate enabled to frame the problem within a control context, ultimately allowing the system to function as a sensor for reconstructing the substrate profile based on the body's deformation.
A continuum manipulator featuring triangular notches was introduced in  Ref. \cite{tim5} for potential medical applications, utilizing wires embedded within bilaterally symmetrical channels for planar actuation. This research emphasizes a mechanics-based kinematic model of the continuum manipulator, employing Timoshenko beam theory to correlate the applied load with the manipulator's configuration. The model segmented the continuum manipulator into multiple V-shaped units, each comprising two 2-node Timoshenko beam elements.

Our research introduces a control strategy that combines Timoshenko modelling with a Sliding Mode Controller (SMC) to regulate the movements of a robotic soft continuum wrist. This integration capitalizes on the strengths of both the approaches. 
The Timoshenko modeling approach offers several significant advantages for the analyses and simulations of soft continuum robots. This method accounts for both shear deformation and rotational effects, which are particularly relevant in soft robotics where materials exhibit considerable flexibility and compliance. By incorporating these factors, the Timoshenko model provides a more accurate representation of the mechanical behavior of soft structures, enabling better predictions of their performance under various loading conditions. Furthermore, this approach facilitates the design and optimization of soft continuum robots to explore a wider range of configurations and material properties, ultimately leading to enhanced functionality and adaptability in complex environments.

The modelling procedures \cite{mod,mod1}, sensor integration \cite{mod2,mod3}, and implementation of control strategies \cite{mpc,mod4} to manage the movements of robotic systems is crucial for enhancing the functionality and user experience of these sophisticated devices \cite{pros}. 
Implementation of advanced control algorithms in robotic systems enables users to execute a broader array of tasks with increased accuracy. 
The effectiveness of an SMC strategy for managing a soft pneumatic actuator was assessed in  Ref. \cite{Skorina}. This control approach enabled the robot to accurately follow dynamic trajectories at varying frequencies while mitigating the impact of external forces, resulting in lower error rates and reduced chattering. In another investigation given in Ref. \cite{khan}, a soft robot was regulated using an SMC that incorporated a PID sliding surface. This controller utilized feedback error to define the PID sliding surface, while a nonlinear SMC ensured the system's adherence to this surface. Experimental results demonstrated that this controller significantly diminished vibration amplitude. Additionally, an SMC framework for a tendon-driven soft robot based on Cosserat rod theory, employing a generalized $\alpha$ method to decrease computational demands during analysis was proposed in  Ref. \cite{Mousa}. The application of an Timoshenko approach in conjunction with SMC offers notable benefits that improve both performance and flexibility across diverse control scenarios. The incorporation of Timoshenko beam theory facilitates the generation of an accurate dynamic model, thereby enabling effective modeling of system dynamics that traditional techniques struggle to showcase \cite{Al}. This feature enhances the controller's ability to adjust to varying system conditions and uncertainties, resulting in greater robustness and stability. Additionally, the synergy of these two strategies can enhance convergence rates and mitigate chattering effects, which are prevalent issues in standard SMC implementations.

A comprehensive review of the literature concerning the development of robotic soft tendon-driven wrist joints indicated that current control strategies for these robots exhibit several shortcomings, including suboptimal control methods, higher computational demands, inaccurate modeling techniques, and non-ideal tuning parameters. This paper presents the methodologies and strategies employed to create a control framework to enhance the motion control of a robotic wrist, achieving greater accuracy while minimizing computational load. The primary contributions of this work are outlined as follows:

\begin{itemize}
    \item Modelling of a soft continuum robotic wrist based on Timoshenko beam theory approach.
    \item Development of a model-based SMC for improving the performance of wrist motions.
    \item A comparison study of the proposed controller with other controllers to demonstrate the advantages of the proposed controller.
    \item An experimental validation proving the effectiveness of the proposed controller during real-time implementations with reduced computational effort.
\end{itemize}

The outline of this paper is given as follows: The components of the wrist and modeling procedures of the wrist using Timoshenko beam theory are given in Section 2. SMC control strategy implemented on the wrist using a dynamic model is demonstrated in Section 3. Section 4 presents the simulation, comparison and experimental results. Discussion and conclusion of the work are given in Sections 5 and 6 respectively.

 \section{Design and Modelling of the Soft Continuum Wrist}
This section discusses about the components and mechanical design of the robotic wrist. It also includes modelings of the wrist utilizing the Timoshenko beam theory method.

\subsection{Design of the Wrist}
\begin{figure}[hbt!]
\centerline{\includegraphics[width=0.65\textwidth, height = 4 in]{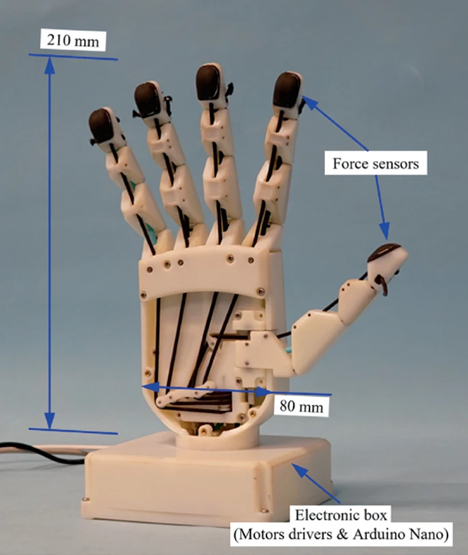}}
\caption{A humanoid hand (PRISMA HAND II)}
\label{hand}
\end{figure}

\begin{figure*}[hbt!]
\centerline{\includegraphics[width=1\textwidth, height = 2.8 in]{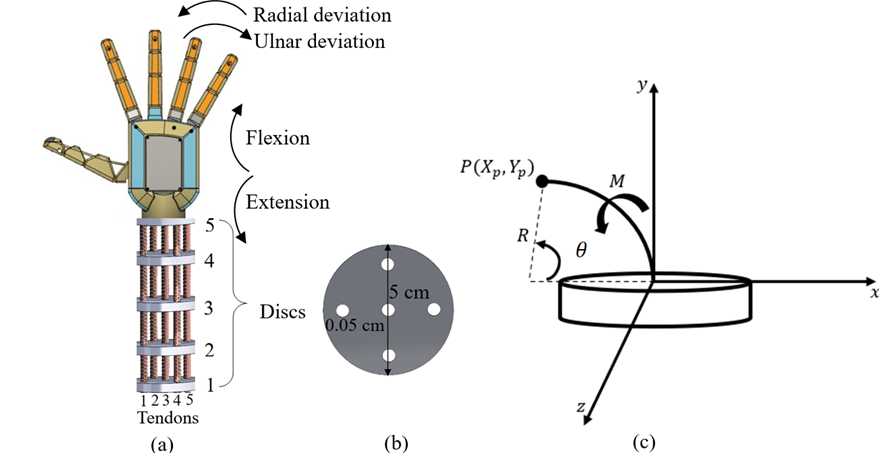}}
\caption{Wrist section (a) conceptual model (b) fabricated model (c) bending configuration }
\label{conceptual model}
\end{figure*}

A soft continuum wrist segment given in Ref. \cite{conf} was designed to support a humanoid hand known as 'PRISMA HAND II' \cite{ref2} (shown in figure \ref{hand}). 
The wrist's design was completed by considering its load-bearing capacity and its ability to move in ulnar, radial, flexion, and extension directions. The integration of the wrist with the PRISMA HAND II, along with the corresponding motion directions, is illustrated in figure \ref{conceptual model}(a). The dimensions of the disc are shown in figure given in Ref. \ref{conceptual model}(b). The wrist assembly consisted of five rigid discs, five springs, and five tendons inserted through the springs. The tendons were adjusted to achieve the required movements in various directions. Tendons 1 and 2 were activated to facilitate radial deviation within the wrist area, whereas tendons 4 and 5 were tasked with enabling movements toward the ulnar side. Furthermore, tendons 1 and 4 played a crucial role in governing extension actions, while tendons 2 and 5 were responsible for managing flexion actions.
The configuration of these components allowed the wrist section to both compress and extend along the axial direction. With respect to the fifth disc, the wrist's motion range was established from -50$^{\circ}$  to +50$^{\circ}$  in all directions, which was sufficient for the PRISMA HAND II to execute manipulation tasks effectively.

\begin{figure*}[hbt!]
\centerline{\includegraphics[width=0.5\textwidth]{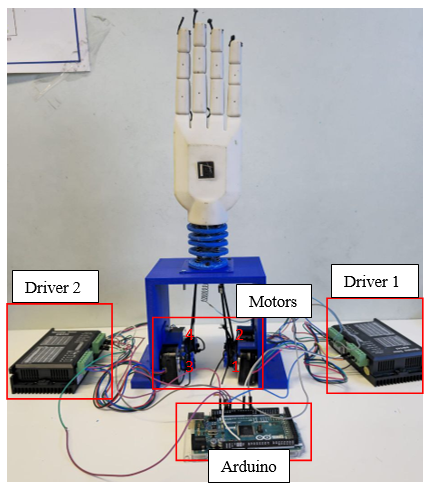}}
\caption{Experimentation setup with the fabricated model}
\label{fabricated}
\end{figure*}

The fabricated wrist section attached with the anthropomorphic robot hand is shown in figure \ref{fabricated}. To achieve ulnar wrist rotation, actuating motors 3 and 4 were engaged in an anticlockwise direction to pull tendons 4 and 5. An ArUco marker affixed to the wrist facilitated the tracking of disc 5's movements during the experiments. Inputs forces for the TMB were provided as desired motion values, which subsequently generated desired bending angles that were fed into the SMC. The SMC produced desired actuation forces, which were then converted into motor torque values and subsequently transformed into PWM signals. These PWM signals were sent to an Arduino Mega 2560, which was connected to two stepper motors via a microstep driver (DM542). To enhance the process, an Arduino module was integrated with Simulink for the transmission of calculated input signals, while a ROS module was utilized to communicate the readings from the ArUco markers into the Simulink environment. 

\subsection{Modelling of the Wrist using Timoshenko Beam Theory Approach}

The modeling of a soft continuum wrist can be effectively approached through the application of Timoshenko beam theory. This theoretical framework provides a comprehensive understanding of the mechanical behavior of the wrist, taking into account both shear deformation and rotational effects, which are particularly relevant in soft materials. By utilizing this theory, one can achieve a more accurate representation of the dynamic and static responses of the wrist, facilitating advancements in the design and control of soft robotic systems.

Variations in tendon tensions result in distinct bending moments on the soft wrist segment, enabling its behavior to be represented as a cantilever beam under a bending moment. The placement of the end effector concerning the wrist's curvature is established based on the Timoshenko beam theory, as cited in Ref. \cite{tim1}. Figure \ref{conceptual model}(c) illustrates the bending configuration of the soft wrist segment, identified by a length $L$, influenced by an anti-clockwise moment, $M$. The location of the hand, denoted as $\bm{P(X_p,Y_p)}$ in the two-dimensional Cartesian plane, along with bending angle $\theta$, is established through the application of equations \ref{eq.1} - \ref{eq.2}.
\begin{equation}
    X_p=R~sin~\theta
    \label{eq.1}
    \end{equation}
    \begin{equation}
      Y_p=R~(1-cos~\theta)
     \label{eq.2}
    \end{equation}
\vspace{0.01 em}

\noindent
The variable $R$ denotes the radius of curvature of the soft wrist section. The overall equation of a beam, derived from quasi-static Timoshenko beam theory given in Ref. \cite{quasi}, incorporating Young's Modulus, $E$, Moment of Inertia, $I$, and deflection, $y$ is given in equation \ref{timo}.
\begin{equation}
    EI{\frac{d^4y}{dx^4} = q(x) -\frac{EI}{KAG} \frac{d^2q}{dx^2}}
    \label{timo}
\end{equation}
\vspace{0.1 em}

\noindent
In this context, $q$, $K$, $A$, and $G$ denote the general load, Timoshenko shear coefficient, cross-sectional area, and shear modulus, respectively. The deflection of the wrist section, denoted as $y$, resulting from a concentrated load $F$ applied to the wrist section, is calculated using equation \ref{timoshenko}.
\begin{equation}
    y(x) = \frac{F(L-x)}{KAG} - \frac{Fx}{2EI}(L^2 - \frac{x^2}{3}) + \frac{FL^3}{3EI}
\label{timoshenko}
\end{equation}
The deformation at the endpoint of the wrist section (\textit{x=L}) is derived from equation \ref{w}.
\begin{equation}
    y(x=L) = \frac{8.8FL}{7.8AG}
    \label{w}
\end{equation}
The vertical deflection of the wrist at the tip is calculated using equation \ref{tip}.
\begin{equation}
    y(x) = \frac{ML^2}{2EI} =\frac{FRL^2}{2EI}
    \label{tip}
\end{equation}
The dynamic model of the wrist segment is then developed using the position obtained from the curvature radius and bending moment based on the dynamic Timoshenko beam theory given in Ref. \cite{dynamic} (including time derivative, $\frac{dy}{dt}$) as given in equation \ref{dyn_tim}.
\begin{equation}
 EI\frac{d^4y}{dx^4} + \rho A \frac{d^2y}{dt^2} = F_t(x)
    \label{dyn_tim}   
\end{equation}
where $\rho$ and $F_t$ represent density and tendon forces of the wrist section respectively. Consequently, quasi-static and dynamic models allow for the identification of the end effector's location and the required tension in the wires, thereby enabling the design and implementation of the control system for the wrist segment.
\section{Sliding Mode Control Strategy}

\begin{figure*}[hbt!]
    \centering
\includegraphics[width=1\textwidth]{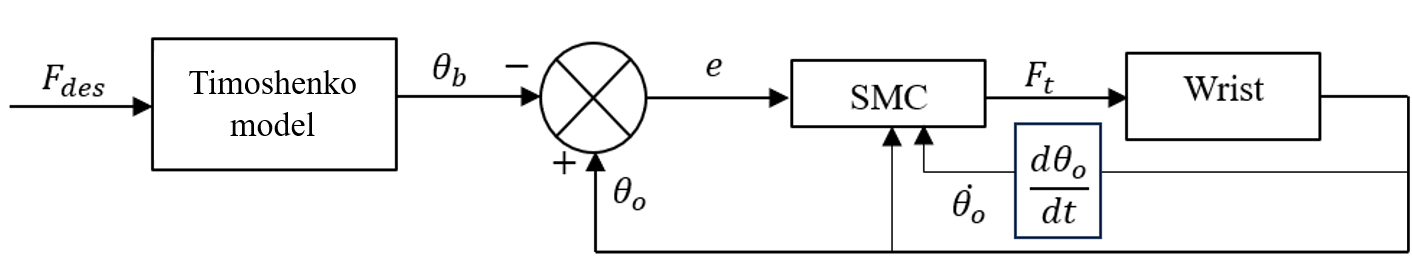}
   \caption{Sliding mode controller strategy}
    \label{hybrid}
\end{figure*}

An SMC strategy has been designed specifically for a soft continuum wrist section, focusing on enhancing its operational efficiency and adaptability. This approach leverages the unique characteristics of soft robotics, allowing for precise control over the robot's movements and interactions with its environment. This control strategy is particularly effective in managing the complexities associated with the non-linear dynamics and uncertainties inherent in soft robotic systems.
By implementing this control strategy, the wrist can effectively navigate complex tasks while maintaining stability and robustness against external disturbances. The proposed controller framework (shown in figure \ref{hybrid}) elevates the strengths of both Timoshenko model and SMC to achieve precise motion control of the soft wrist component.
Desired forces ($F_{des}$) were fed to a Timoshenko Model Block (TMB) and corresponding bending angles, $\theta_b$ were obtained as the output. In TMB, $y_{des}$ were computed based on equation \ref{tip} and desired bending angles, $\theta_b$ were calculated based on equation \ref{theta}.
\begin{equation}
    \theta_{b} = \frac{y_{des}}{L}
    \label{theta}
\end{equation}
Tendon forces, $F_t$ were obtained as the output from SMC block and fed to wrist model. Output bending angles, $\theta_o$ were used for calculating errors, $e(t)$ as given in equation \ref{eq:12}
\begin{equation}
    e(t)=\theta_o -\theta_b
    \label{eq:12}
\end{equation}
$\theta_o$, $\dot \theta_o$, and $e(t)$ were given as the input to SMC block for correcting the output motions. The implementation of the SMC served as an effective strategy for regulating the motions of the soft wrist due to its robustness against uncertainties and external disturbances. %which are common in the operation of soft robots. 
Dividing equation \ref{dyn_tim} by L and eliminating the first element (since we are focusing only on the wrist tip), we get the following equation \ref{ss}
\begin{equation}
   \frac{\rho A \ddot y}{L} = \frac{F_t}{L}
   \label{ss}
\end{equation}
Since $\frac{\ddot y}{L} = \ddot \theta $, equation \ref{ss} can be rewritten as given in equation \ref{ss1}
\begin{equation}
    \rho A \ddot \theta = \frac{F_t}{L}
    \label{ss1}
\end{equation}
Let us consider the following representations given in equation \ref{eq_1}-\ref{eq_3}:
\begin{equation}
\theta_1 =\theta_0
\label{eq_1}
\end{equation}
\begin{equation}
    \dot \theta_1 = \theta_2 = \dot \theta_0
\label{eq_2}
\end{equation}
\vspace{0 em}
\begin{equation}
\ddot \theta_1 = \dot \theta_2 = \ddot \theta_0
\label{eq_3}
\end{equation}
\vspace{0.5 em}

\noindent
The linear state-space form of the equation \ref{ss1} is obtained as given in equation \ref{ssf}
\begin{equation*}
  \dot \theta_1 = \theta_2 
\end{equation*}
\begin{equation}
\dot \theta_2 = \frac{F_t}{\rho A L} 
   \label{ssf}
\end{equation}
\vspace{0.5 em}

\noindent
Let us consider the state of the system given in equation \ref{st}
\begin{equation}
\bm{\theta} =
    \begin{bmatrix}
        { \theta_1} \\
       { \theta_2}
    \end{bmatrix}
    =
    \begin{bmatrix}
       { \theta_0} \\
        {\dot \theta_0}
    \end{bmatrix}
    \label{st}
\end{equation}
Consider the non-linear conceptual form of wrist system given in equation \ref{eq:13}
\begin{equation}
       \bm{\dot \theta} =
    \begin{bmatrix}
        {\dot \theta_0} \\
       { \ddot \theta_0}
    \end{bmatrix} = \bm{f(\theta) + g(\theta)}U, \bm{\theta}\in R^{n}, {U}\in R
    \label{eq:13}
\end{equation}
where  \bm{$f(\theta)$} is the vector field  given in equation \ref{eq:14}
\begin{equation}
\bm{f(\theta)} = 
\begin{bmatrix}
   { \dot \theta_o} \\
   { 0}
    %M^{-1}[J^T F-(C+D)\dot \theta_{o} - K\theta_o - G]
\end{bmatrix}
\label{eq:14}
\end{equation}
\bm{$g(\theta)$} is the function which maps $U$ to the force field of the system as given in equation \ref{eq:15}
\begin{equation}
\bm{g(\theta)} = 
\begin{bmatrix}
    {0} \\
   { \frac{1}{\rho A L}}
\end{bmatrix}
\label{eq:15}
\end{equation}
\vspace{0.5 em}

\noindent
\bm{$f(\theta)$} and \bm{$g(\theta)$} were computed based the state-space form of the equation \ref{ss}. $U$ is the control input (in our case, $F_t$) given in the equation \ref{eq:16}
\begin{equation}
    U = -[U_{equvalent} + U_{switching}]
    \label{eq:16}
\end{equation}
$U_{equivalent}$ is obtained based on the equation \ref{eq:17}
\begin{equation}
    U_{equivalent} = \frac {L_f(\sigma(\theta))}{L_g(\sigma(\theta))}
    \label{eq:17}
\end{equation}
where $L_f(\sigma(\theta)$ and $L_g(\sigma(\theta)$ are the lie derivatives of functions \bm{$f(\theta)$} and \bm{$g(\theta)$} with respect to $\sigma(\theta)$. $\sigma(\theta)$ is the sliding surface function given in equation \ref{eq:18}
\begin{equation}
    \sigma(\theta) = P_1e(t) + P_2\dot \theta_o
    \label{eq:18}
\end{equation}
where $P_1$ and $P_2$ are two tuning parameters to stabilise the controller and reduce the output errors to a tolerance range. The importance of tuning parameters in an SMC cannot be overstated, as these parameters play a crucial role in determining the performance and stability of the control system. Proper tuning of $P_1$ and $P_2$ ensures that the controller can effectively manage system dynamics, particularly in the presence of uncertainties and external disturbances. 
Sliding surface $S$ was selected based on the convergence in finite time to guarantee optimal control and is given in equation \ref{eq:19}
\begin{equation}
    S = { \bm\theta \in R^{n}: \sigma(\theta) = 0}
    \label{eq:19}
\end{equation}
$U_{switching}$ is given in equation \ref{eq:20}
\begin{equation}
    U_{switching} = \frac {1}{L_g(\sigma(\theta))} P_3~tanh(\sigma (\theta))
    \label{eq:20}
\end{equation}
where $P_3$ is a control parameter to increase the stability of the controller. We first implemented $sgn $ function instead of $tanh$ function in equation \ref{eq:20} and compared the performances of the SMC. The performance of the SMC employing $tanh$ function was better in terms of lower computational time. In order to maintain the system to the sliding surface, the following condition is applied as given in equation \ref{eq:21}
\begin{equation}
    U= 
    \begin{Bmatrix}
        U^+, &\sigma(\theta)>0 \\
        U^-, &\sigma(\theta)<0
    \end{Bmatrix}
    \label{eq:21}
\end{equation}

\section{Results and Discussion}
This paper presented the modeling and development of a controller for a wrist component intended to facilitate payload handling and intricate movements of a robotic system. The subsequent section compares the proposed SMC approach with another controller developed for the wrist section. Additionally, this section will feature simulation study and experimental validation of the SMC to assess the controller's performance in real-time applications.

\subsection{Simulation Study}
\begin{figure}[hbt!]
    \centering
\includegraphics[width=0.8\textwidth]{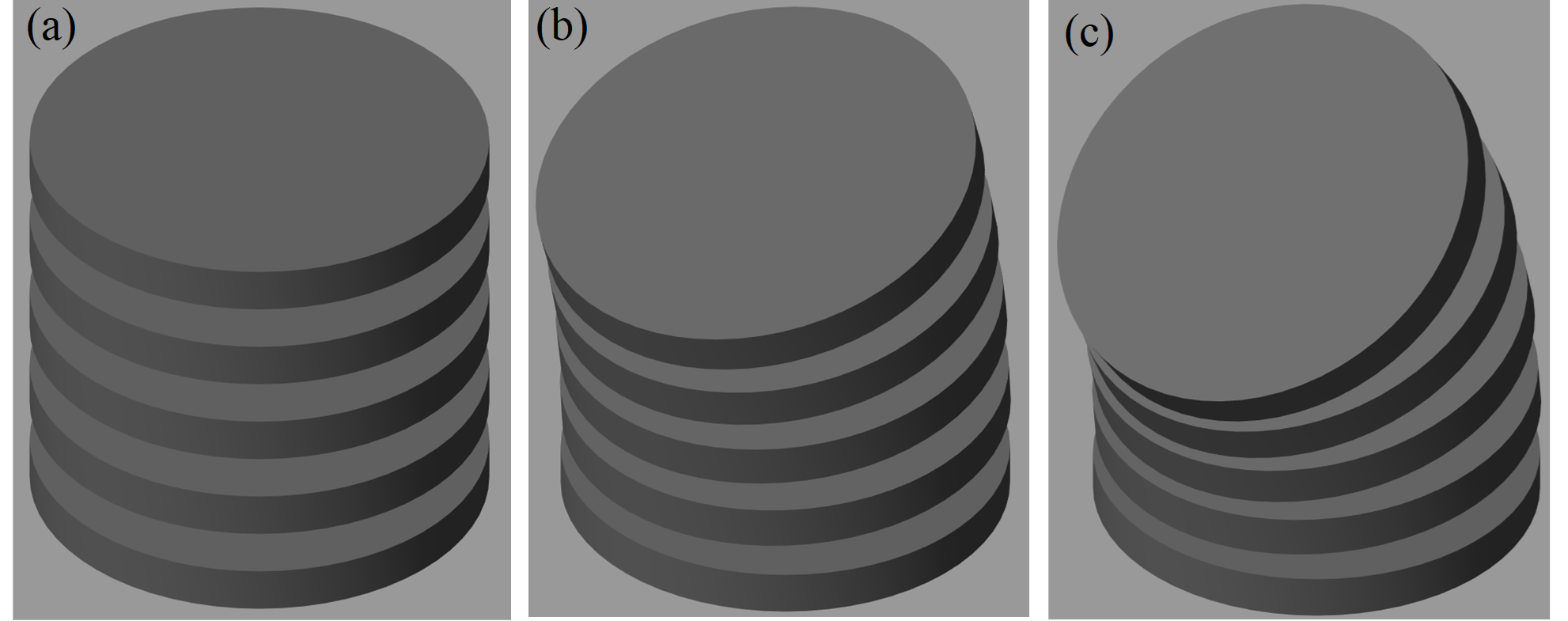}
   \caption{Motion of wrist in ulnar deviation direction (Rearview)}
    \label{Motion of wrist}
\end{figure}

Controller loops were developed within Simulink, a software platform based on MATLAB, and executed on a personal computer equipped with an Intel Core i7 processor and 16 GB of RAM. The wrist segment is capable of moving along trajectories in the directions of radial deviation, ulnar deviation, flexion, and extension. The motions of wrist along ulnar deviation are shown in figures \ref{Motion of wrist} (a) - (c).

The wrist segment was modeled to bend from its initial position, as depicted in figure \ref{Motion of wrist}, to a final angle of $30^{0}$ in ulnar direction relative to the last notable point on disc 5 connected to the hand. 
This research focused solely on the system's responses during motion in the ulnar deviation direction, as all movements were symmetrical, enabling the extrapolation of responses in other directions from the results obtained. The damping and stiffness coefficients of the wrist were established as $0.615$ Nm and $0.105$ Nms, respectively. By adeptly addressing the non-linearities and uncertainties inherent in the soft wrist's motions, SMC contributed to smoother and more efficient movements, enhancing the system's overall performance. The tuning parameters ($P_1$, $P_2$, and $P_3$) were crucial for balancing system responsiveness and stability and initially set to 1. These parameters were optimized using Particle Swarm Optimization (PSO) to minimize output error and chattering, resulting in final values of 50, 1, and 40 for $P_1$, $P_2$, and $P_3$, respectively.

\begin{figure}
\centerline{\includegraphics[width=0.8\textwidth, height = 2.9 in]{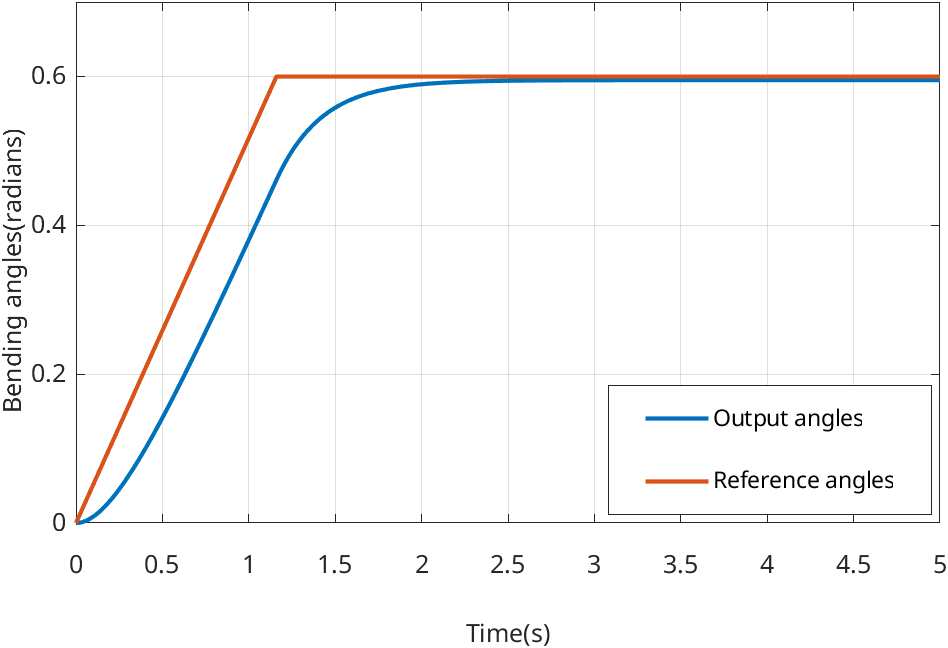}}
\caption{ 
Comparison between the desired and the output bending angles } 
\label{error_2_new}
\end{figure}

\begin{figure}
\centerline{\includegraphics[width=0.8\textwidth, height = 2.6 in]{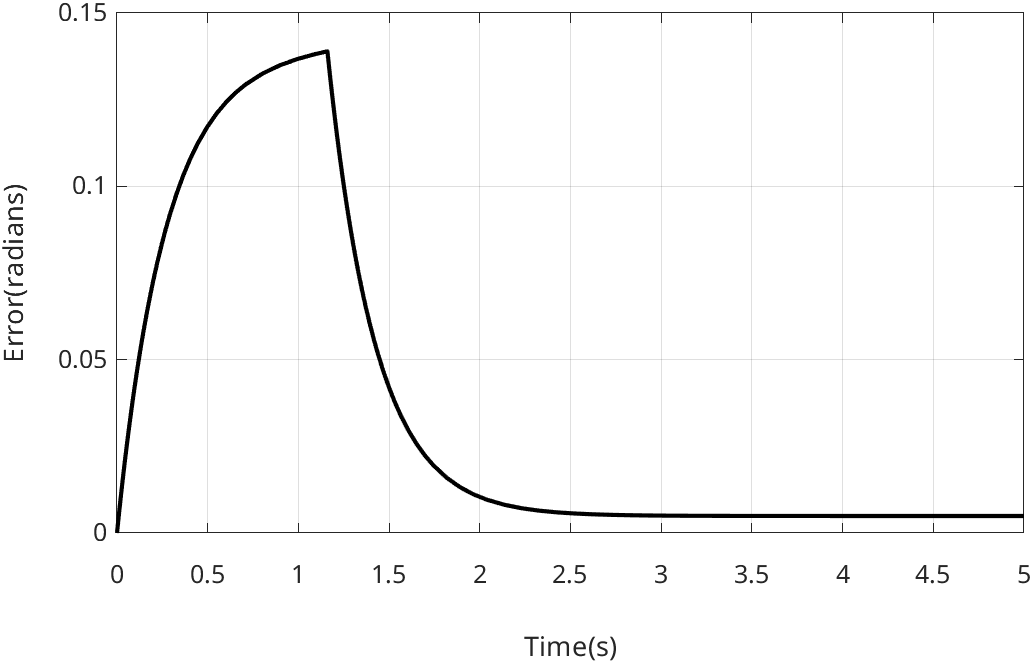}}
\caption{ 
Error in angles with respect to reference bending angles using SMC during ulnar deviation} 
\label{error_3}
\end{figure}

\begin{figure}
\centerline{\includegraphics[width=0.9\textwidth, height = 2.7 in]{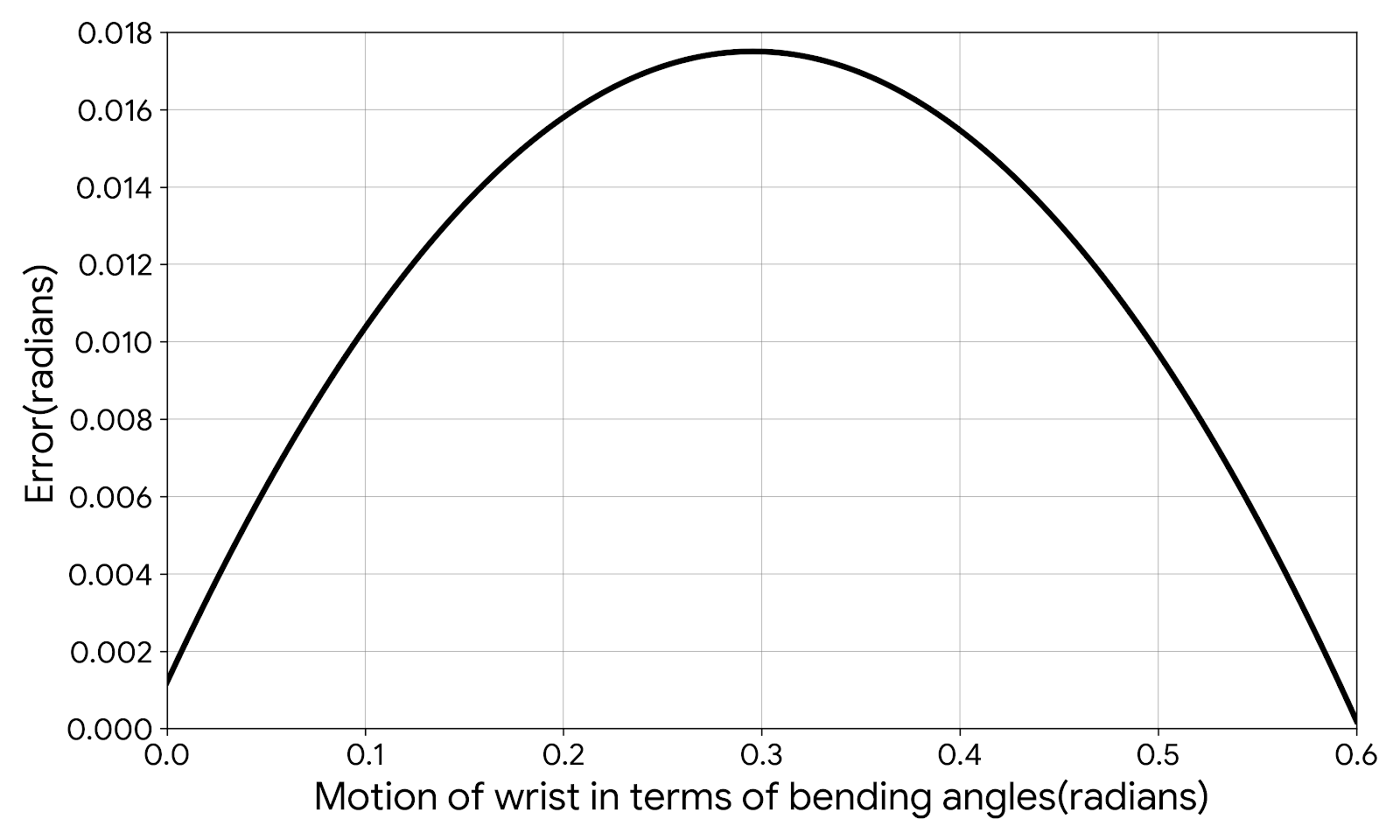}}
\caption{ 
Error in angles with respect to bending angles using SMC during ulnar deviation} 
\label{error_3_new}
\end{figure}

The comparison between the desired and the output bending angles achieved through the SMC strategy during the simulation study is illustrated in figure \ref{error_2_new}. 
%\begin{figure}[hbt!]
%\centerline{\includegraphics[width=1\textwidth, height = 1.7 in]{Figures_5_6.png}}
%\caption{Comparison of desired and output bending angles obtained using SMC during ulnar deviation} 
%\label{error}
%\end{figure}
\begin{comment}
\begin{figure}[hbt!]
\centerline{\includegraphics[width=0.9\textwidth, height = 2.6 in]{angles.png}}
\caption{Comparison of desired and output bending angles obtained using SMC during ulnar deviation} 
\label{error}
\end{figure}
\end{comment}
The errors in the desired and output bending angles are showcased in figure \ref{error_3}. The error with respect to the motion of wrist from 0 to 0.6 radians are shown in figure \ref{error_3_new}.
The Root Mean Square Error (RMSE) calculated for these discrepancies was obtained as $1.67 \times 10^{-2}$ radians. Additionally, the settling time and steady-state error were measured as $1.9$ seconds and $3.73 \times 10^{-3}$ radians, respectively. All these values, including the RMSE, settling time, and steady-state error, were found to be within the acceptable tolerance limits.
%Figure:error smc
\begin{comment}
\begin{figure}
\centerline{\includegraphics[width=0.9\textwidth, height = 2.8 in]{Error_simulations.png}}
\caption{ Error in angles with respect to time using SMC during ulnar deviation} 
\label{error_1}
\end{figure}
\end{comment}
%Figure:error smc


\begin{figure}
\centerline{\includegraphics[width=0.8\textwidth, height = 3.4 in]{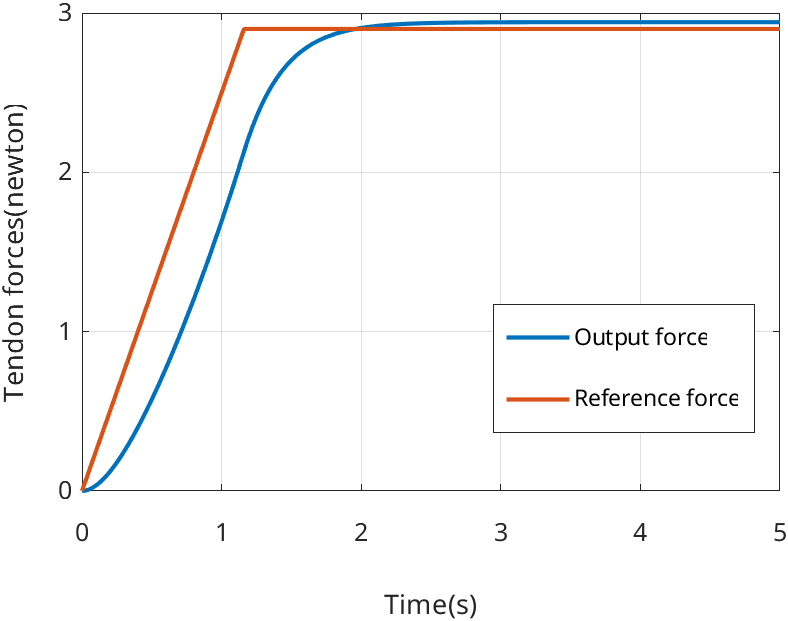}}
\caption{ 
Comparison of reference and output tendon forces} 
\label{error_force}
\end{figure}

\begin{comment}
\begin{figure}
\centerline{\includegraphics[width=1\textwidth, height = 1.7 in]{Figures_7_8.png}}
\caption{ 
Error in angles with respect to bending angles using SMC during ulnar deviation} 
\label{error_2}
\end{figure}
\end{comment}
The comparison of reference and output tendon forces is shown in figure \ref{error_force}.
\begin{comment}
\begin{figure}[hbt!]
\centerline{\includegraphics[width=1\textwidth, height = 3.3 in]{force.png}}
\caption{ Comparison of desired and output tendon forces using SMC during ulnar deviation} 
\label{error_3}
\end{figure}
\end{comment}
The figure depicts the nature of output force generated during the simulation study in comparison with reference values. The output converges along with the reference force values around $1.9$ seconds.

The performance of the SMC was compared with other controllers developed for the wrist section. RMSE, settling time and  steady state error were compared to determine a better controller for the desired task as given in table \ref{table:2}. Comparison of the proposed SMC with a geometric variable strain controller (GVSC) and a conventional PID based controller developed for the wrist for controlling motion of wrist in ulnar deviation direction is given in Table \ref{table:2}. 
\begin{table}[hbt!]
\caption{Comparison study}
  \centering
    \begin{tabular}{c c c c}
    \hline 
    Parameters  &GVSC &PID  &SMC \\
    \hline
     RMSE (Radians)   &0.029 &0.266 &0.016 \\
    Settling time (Seconds)   &3.180 & 5.740 &1.900   \\
    Steady state error (Radians)  &0.014 &1.210 &0.003     \\
    \hline 
     \end{tabular}
  \label{table:2}
\end{table}
The kinematic and dynamic modelings of the wrist section were conducted using the GVSC approach based on Cosserat rod theory and a generalized coordinate method, respectively. The 
 dynamic model was used for designing a force control scheme for controlling motions of the wrist section. Additionally, a PID controller based on kinematic model determined employing GVSC approach was also developed for the wrist section. The outputs obtained from these controllers are showcased in Table \ref{table:2}. The GVSC utilizes a force-modulation strategy, carefully adjusting strain distribution along the wrist’s structure. By dynamically varying strain forces, GVSC enhances the system’s ability to perform dexterous wrist movements, leading to improved precision and responsiveness. This approach is particularly beneficial in accommodating external disturbances and unexpected variations in load. In contrast, the SMC follows a gain-related adaptation method, which systematically adjusts controller gains based on real-time system feedback. This ensures robust stability by minimizing deviations from the desired trajectory while simultaneously optimizing computational efficiency. Unlike traditional control methods, SMC leverages its adaptive gain mechanism to effectively handle nonlinearities and uncertainties in wrist motion dynamics, resulting in smoother transitions, reduced errors, and faster settling times. Additionally, its lower computational demand makes it well-suited for real-time applications, where processing speed and accuracy are crucial. The PID controller is a linear control strategy, meaning it struggles to handle nonlinear and dynamic variations in wrist joint movement. In contrast, SMC and GVSC incorporate adaptive mechanisms and SMC utilizes gain adaptation, while GVSC employs strain-based force modulation allowing them to better adjust to variations in motion dynamics. Due to its fixed-gain structure, PID controllers may struggle with persistent errors, especially in highly dynamic environments. SMC’s sliding mode properties actively correct errors by adapting control gains, leading to improved accuracy in tracking wrist joint movements. Hence, the proposed SMC scheme was found to be more efficient in controlling the given task with fewer errors by using lower computational efforts. The RMSE, settling time and steady state error obtained using SMC were found to be lower as evident from the table \ref{table:2}. The RMSE of SMC was obtained as 0.016 radians whereas RMSE values of GVSC and PID controller are 0.029 and 0.266 respectively. Settling times of GVSC and PID controller were obtained as 3.180 and 5.740 seconds respectively. SMC showcased a settling time of 1.9 seconds lower than other two controllers. The steady state error from SMC was obtained as 0.003 radians where as GVSC showcased a value of 0.014 radians and PID controller showcased 1.21 radians. PID controller performed least efficient in terms of all parameters compared to the other two controllers.

\subsection{Experimental Validation of SMC}
\begin{figure*}[hbt!]
\centerline{\includegraphics[width=1\textwidth,height = 2.5 in]{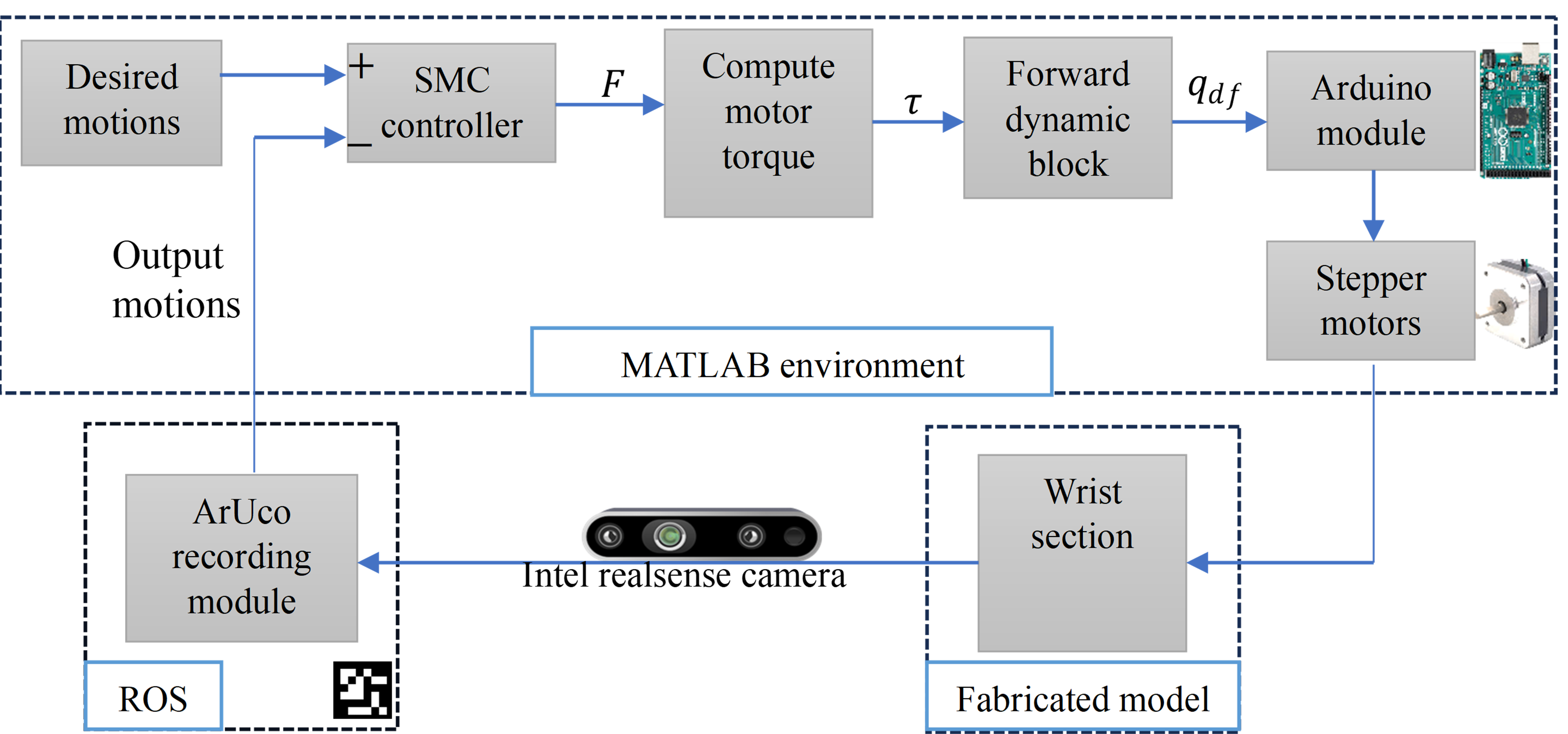}}
\caption{ Flowchart of experimentation} 
\label{flow}
\end{figure*}

\begin{figure*}[hbt!]
\centerline{\includegraphics[width=0.8\textwidth,height = 2.5 in]{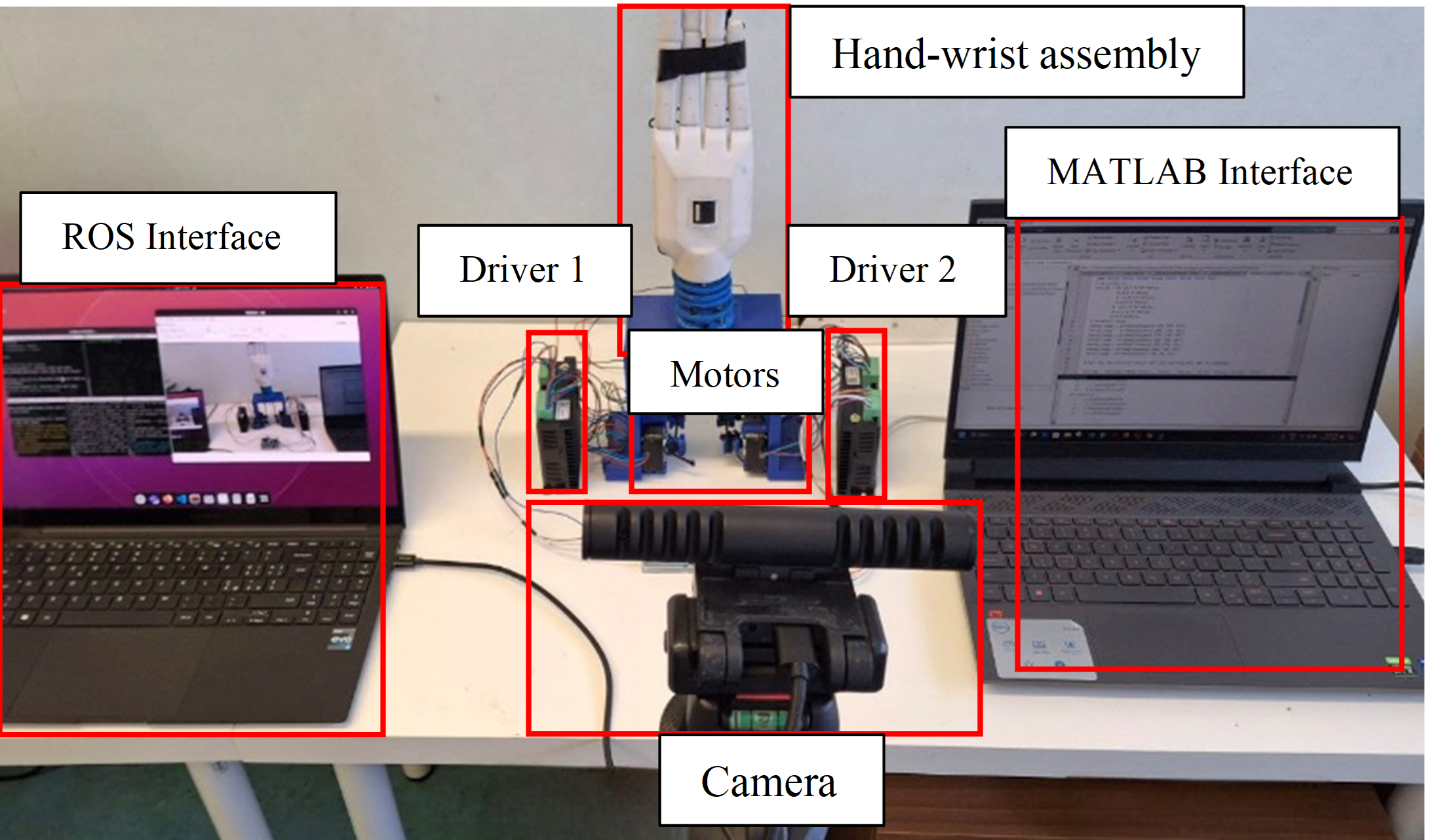}}
\caption{ Experimentation set up including all the components} 
\label{exp_setup}
\end{figure*}

\begin{figure*}
\centerline{\includegraphics[width=0.9\textwidth, height = 1.2 in]{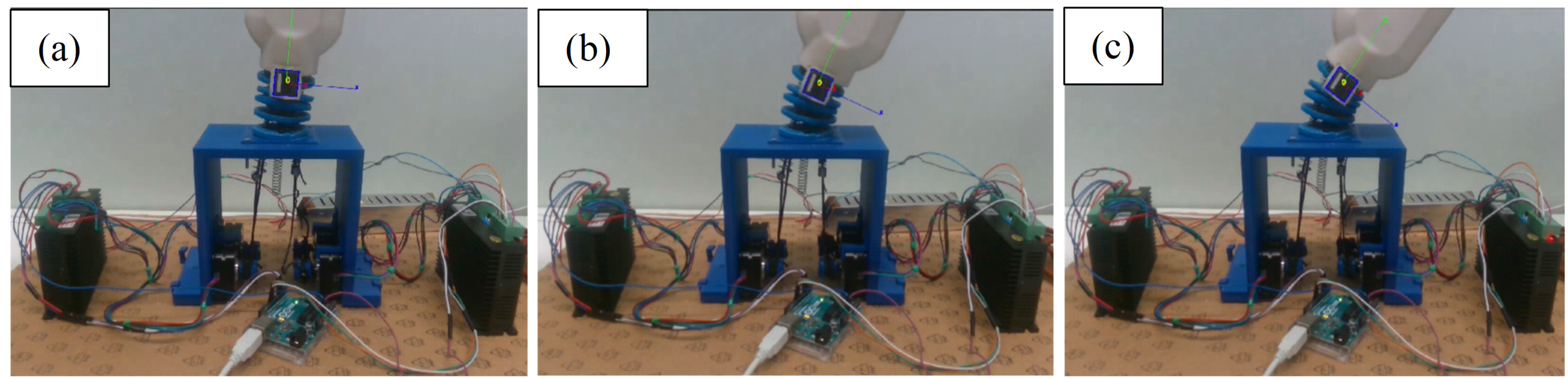}}
\caption{Motion of hand in ulnar deviation direction during experimentation} 
\label{exp}
\end{figure*}

\begin{figure}[hbt!]
\centerline{\includegraphics[width=0.9\textwidth, height = 2.9 in]{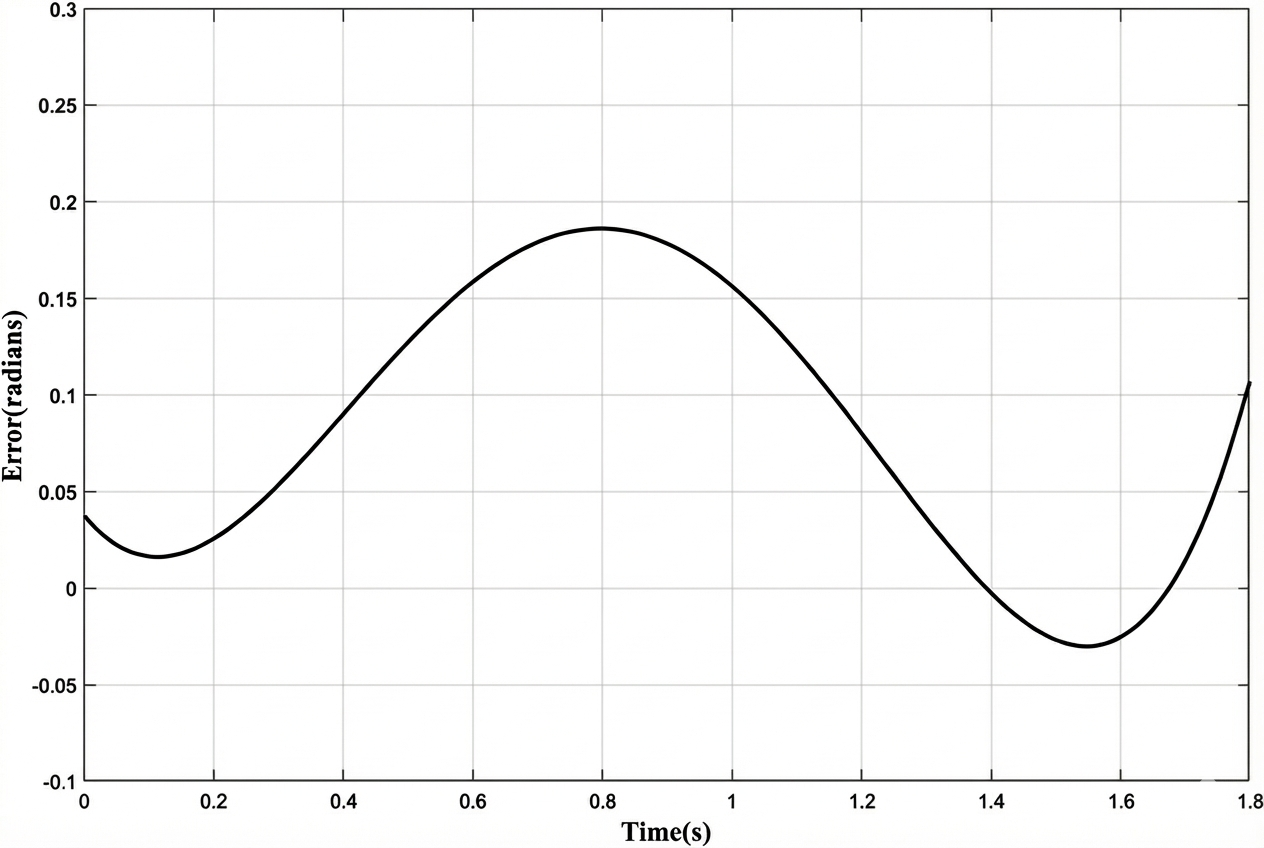}}
\caption{ Errors in the hand motion during experimentation} 
\label{error_exp_new}
\end{figure}

\begin{figure}[hbt!]
\centerline{\includegraphics[width=0.9\textwidth, height = 2.8 in]{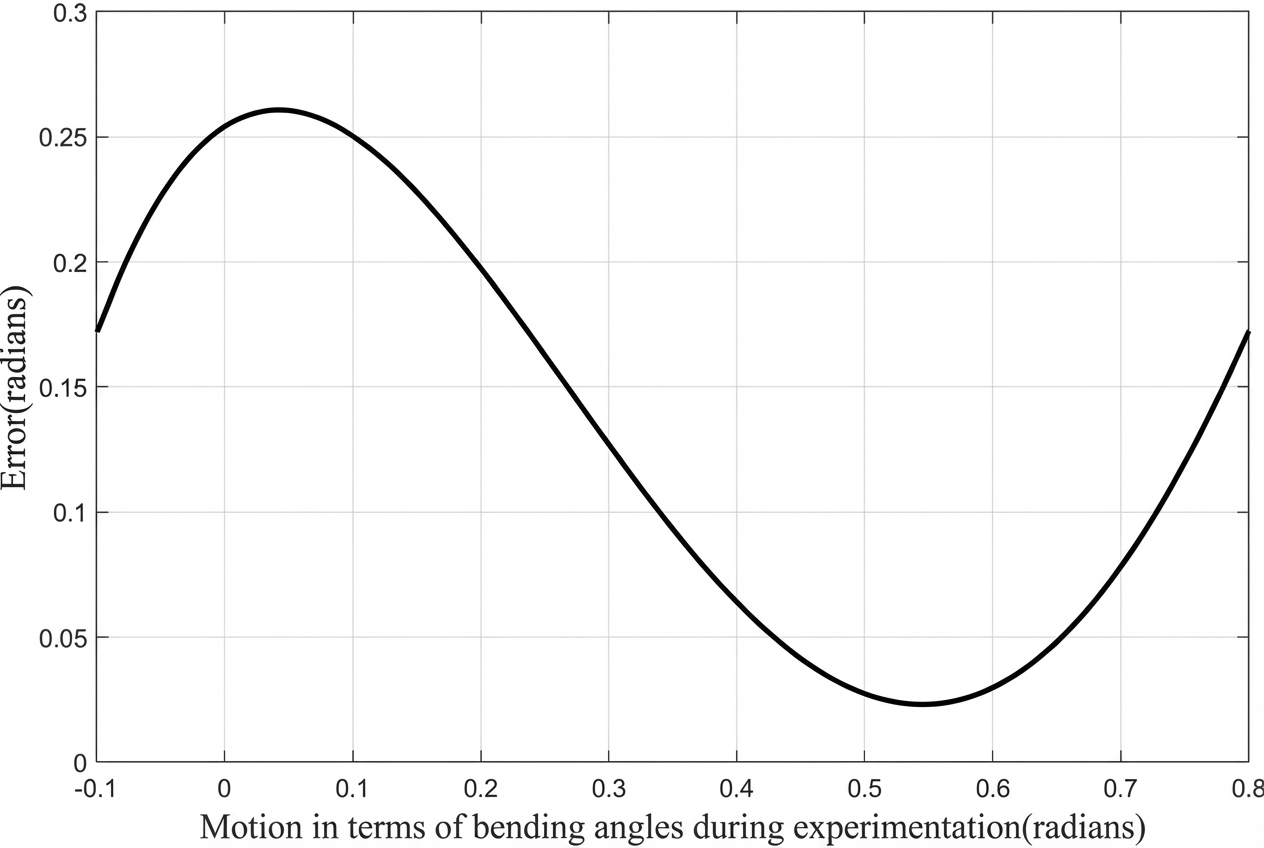}}
\caption{ Errors in the hand motion in terms of angles during experimentation} 
\label{error_exp}
\end{figure}

\begin{figure}[hbt!]
\centerline{\includegraphics[width=0.9\textwidth, height = 3 in]{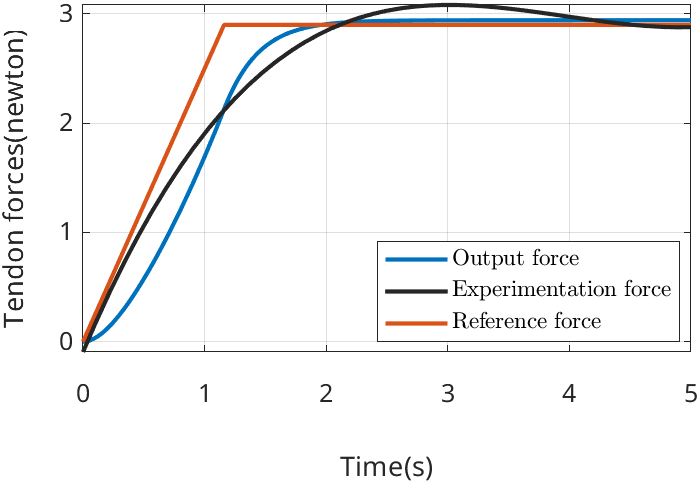}}
\caption{ Comparison of force outputs during simulation and experimentation with reference force values} 
\label{error_exp1}
\end{figure}

\begin{figure}[hbt!]
\centerline{\includegraphics[width=0.9\textwidth, height = 3 in]{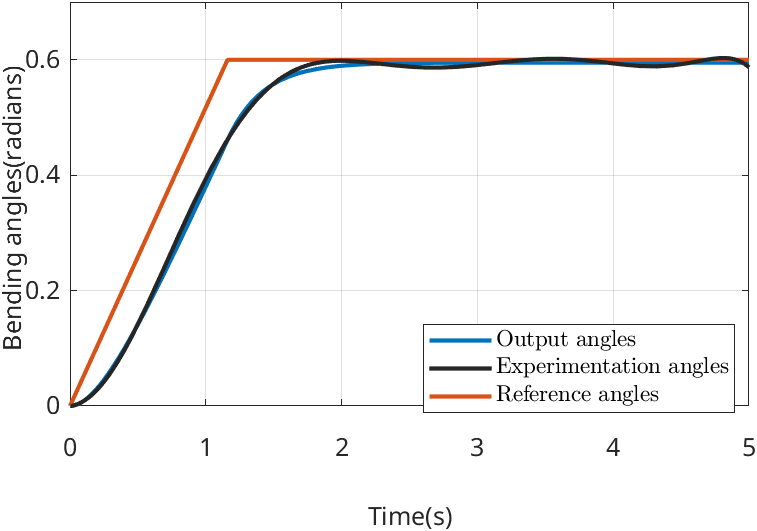}}
\caption{ Comparison of bending angle outputs during simulation and experimentation with reference bending angle values} 
\label{error_exp2}
\end{figure}

The performance of the proposed controller strategy was validated using a fabricated wrist section attached to the PRISMA hand II, as illustrated in figure \ref{fabricated}. %To achieve ulnar wrist rotation, actuating motors 3 and 4 were engaged in an anticlockwise direction to pull tendons 4 and 5. An ArUco marker affixed to the wrist facilitated the tracking of disc 5's movements during the experiments. Inputs forces for the TMB were provided as desired motion values, which subsequently generated desired bending angles that were fed into the SMC. The SMC produced desired actuation forces, which were then converted into motor torque values and subsequently transformed into PWM signals. These PWM signals were sent to an Arduino Mega 2560, which was connected to two stepper motors via a microstep driver (DM542). To enhance the process, an Arduino module was integrated with Simulink for the transmission of calculated input signals, while a ROS module was utilized to communicate the readings from the ArUco markers into the Simulink environment. 
The flowchart of the experimentation is shown in figure \ref{flow} and experimentation environment is shown in figure \ref{exp_setup}.
The movements of the wrist that facilitate robotic system motions during the experiment are illustrated in figures \ref{exp} (a) - (c).
Figures \ref{error_exp_new} and \ref{error_exp} illustrate the errors observed during hand motions with respect to time and bending angles of wrist respectively. The experimental results yielded RMSE values for bending angles, settling time, and steady-state error of 0.2 radians, 2.95 seconds, and 0.02 radians, respectively. 
%\begin{figure}[hbt!]
%\centerline{\includegraphics[width=1\textwidth, height = 1.7 in]{Figures_13_14.png}}
%\caption{ Errors in the hand motion during experimentation} 
%\label{error_exp_new1}
%\end{figure}
Figures \ref{error_exp1} and \ref{error_exp2} showcase the comparison of tendon forces and bending angles during simulation and experimentation with reference values. 
\begin{comment}
\begin{figure}[hbt!]
\centerline{\includegraphics[width=1\textwidth, height = 3.2 in]{error_angles_exp.png}}
\caption{ \hl {Errors in the hand motion during experimentation in terms of bending angles} }
\label{error_exp1}
\end{figure}
\end{comment}
The data distinctly showed that the error values noted in the experimental phase were higher than those documented in the simulation study. The main reason for the elevated error margin during experimentation was the diminished stiffness of the springs used in the wrist segment.
\section{Discussion}
Our study presented an approach to enhance the movements of a tendon-driven robotic wrist joint using a SMC. The proposed wrist joint improved precision and adaptability in motion, offering greater control and accuracy in trajectory tracking. The development of an efficient SMC introduced a robust control mechanism that can be applied to robotic systems, where the ability to achieve fast dynamic responses while maintaining stability has broader implications for robotics and automation. By focusing on adaptability and precision, the study contributed to ongoing efforts to improve dexterous robotic systems for complex manipulation tasks.
The Timoshenko-based modeling approach proved effective in characterizing the kinematic and dynamic properties of the wrist joint. However, it simplified certain biomechanical and mechanical factors. Real-world robotic systems often encounter complex nonlinearities, frictional effects, and compliance interactions that are difficult to fully capture in tendon-driven mechanisms. Although the study highlighted that SMC is optimized for computational efficiency, practical implementations may face challenges such as chattering. If the model is not sufficiently fine-tuned, excessive control effort may be required to minimize instability. In this work, parameters were carefully adjusted and a tanh function clipped to a control signal range was employed to mitigate chattering effects.
While simulation studies provided valuable insights, they may not fully account for mechanical constraints such as backlash, friction, and wear-and-tear. Experimental evaluations demonstrated promising results demonstrating stable motions. The current design is tailored to a specific wrist joint mechanism, and future iterations require additional adaptability features to accommodate varying robotic architectures and application-specific requirements.
A critical aspect of advancing robotic wrist mechanisms lies not only in technical accuracy but also in long-term usability and integration into broader robotic platforms. Incorporating extended experimental trials, robustness testing, and feedback from practical deployments will provide deeper insights into system performance and reliability. Future implementations will also explore adaptive control strategies to further personalize motion and functionality, as well as compact actuation solutions to improve efficiency and integration.

Although the tuning parameters $P_1$, $P_2$,and $P_3$ play a key role in shaping the controller’s responsiveness, the SMC framework is intrinsically robust to bounded variations in system parameters. During the full duration of experimental testing, the wrist mechanism exhibited typical changes such as minor tendon slackening, frictional variation, and spring compliance drift. The controller maintained stable convergence without requiring re‑tuning, indicating that the SMC structure effectively compensates for gradual parameter changes. While a long‑term degradation study is beyond the scope of this work, the observed experimental stability suggested that the proposed controller can tolerate moderate variations in mechanical characteristics over time.

While the proposed control strategy demonstrates strong performance on the current PRISMA HAND II wrist–hand assembly, the present hardware configuration is not optimized for portability. Future work will focus on redesigning the actuation and electronics layout to create a more compact and self‑contained module. Enhancing portability will enable the system to be deployed in unstructured or uncertain external environments, allowing us to investigate interaction dynamics, environmental disturbances, and real‑world manipulation tasks. Such studies will further validate the robustness of the Timoshenko‑based model and the proposed SMC framework under more challenging operating conditions.

\section{Conclusion}
The establishment of an advanced control framework is crucial for enhancing the functionality of robotic wrist mechanisms. This paper introduced a Timoshenko model-based SMC system capable of achieving dynamic and precise control of a tendon-driven robotic wrist. The integration of the Timoshenko modeling strategy with SMC provided notable benefits, improving both system performance and resilience. By leveraging the robustness of SMC, the system was able to sustain intended motion trajectories despite fluctuations in the robot’s physical characteristics and varying operating conditions. Furthermore, the combination of Timoshenko modeling and SMC optimized controller efficiency by reducing computational effort while maintaining accuracy.
The simulation results indicated that SMC achieved a lower RMSE compared to other similiar controllers, with reduced settling times and steady-state errors. However, during practical experiments, the RMSE values were found to be higher than those in simulations, attributed to variations in spring stiffness.  Comparison between simulation and experimental results revealed higher error values in the experimental phase, with RMSE for bending angles increasing from $1.67 \times 10^{-2}$ radians in simulation to $0.2$ radians experimentally. Furthermore, the experimental settling time and steady state error were recorded as $2.95 $ seconds and $0.02$ radians, respectively, compared to $1.9$ seconds and $3.73 \times 10^{-3}$ radians in the simulation. These discrepancies highlight the challenges in translating theoretical performance to real-world implementation, emphasizing the need for further refinement in wrist control systems. The findings demonstrate that the proposed framework offers a reliable solution for trajectory tracking and dynamic response in robotic joints. While the current study focused on mechanism design and control validation, future work may explore compact actuation strategies, adaptive control methods, and extended experimental evaluations to strengthen applicability across diverse robotic platforms.
Future research will aim to redesign the wrist for improved structural integrity and refine control strategies by incorporating real-time sensor feedback to improve motion accuracy.

\section*{Declarations}
Ethics approval and consent to participate: Not applicable
\newline
Consent for publication: Not applicable
\newline
Availability of data and material: Data and explanations related to this study are available
upon reasonable request by contacting the corresponding author.
\newline
Funding:
This work was supported by the Italian Ministry of Research
under the complementary actions to the NRRP ”Fit4MedRob - Fit
for Medical Robotics” Grant (PNC0000007). 
\newline
Conflict of Interest:  The authors declare that they have no conflict of interest.
\newline        
Authors' contributions: Shifa Sulaiman conceptualized the study, developed the methodology, and wrote the manuscript. Mohammad Gohari contributed to the conceptualization of the study. Shifa Sulaiman and Francesco Schetter conducted the simulation study and experimental validations. Fanny Ficuciello acquired funding and supervised the research. All authors have read and approved the final manuscript.


\begin{thebibliography}{99}
 \bibitem{soft}
 Nguyen, T. D., \& Burgner-Kahrs, J. (2015, September). A tendon-driven continuum robot with extensible sections. In 2015 IEEE/RSJ International Conference on Intelligent Robots and Systems (IROS) (pp. 2130-2135). IEEE.


 \bibitem{soft1}
 Li, M., Kang, R., Geng, S., \& Guglielmino, E. (2018). Design and control of a tendon-driven continuum robot. Transactions of the Institute of Measurement and Control, 40(11), 3263-3272.

\bibitem{wockenfuss2022design} Wockenfuß, W. R., Brandt, V., Weisheit, L., \& Drossel, W. G. (2022). Design, modeling and validation of a tendon-driven soft continuum robot for planar motion based on variable stiffness structures. \textit{IEEE Robotics and Automation Letters}, \textit{7}(2), 3985-3991. https://doi.org/10.1109/LRA.2022.3149031

\bibitem{tutcu2021quasi} Tutcu, C., Baydere, B. A., Talas, S. K., \& Samur, E. (2021). Quasi-static modeling of a novel growing soft-continuum robot. \textit{The International Journal of Robotics Research},  \textit{40}(1), 86-98. https://doi.org/10.1177/0278364919893438

\bibitem{escande}Escande, C., Chettibi, T., Merzouki, R., Coelen, V., \& Pathak, P. M. (2014). Kinematic calibration of a multisection bionic manipulator. \textit{IEEE/ASME Transactions on Mechatronics},  \textit{20}(2), 663-674. https://doi.org/10.1109/TMECH.2014.2313741

\bibitem{huang}Huang, X., Zou, J., \& Gu, G. (2021). Kinematic modeling and control of variable curvature soft continuum robots. \textit{IEEE/ASME Transactions on Mechatronics},  \textit{26}(6), 3175-3185. https://doi.org/10.1109/TMECH.2021.3055339

\bibitem{xavier}Xavier, M. S., Fleming, A. J., \& Yong, Y. K. (2021). Finite element modeling of soft fluidic actuators: Overview and recent developments. \textit{Advanced Intelligent Systems},  \textit{3}(2), 2000187.  https://doi.org/10.1002/aisy.202000187

\bibitem{renda}Giorelli, M., Renda, F., Calisti, M., Arienti, A., Ferri, G., \& Laschi, C. (2015). Neural network and Jacobian method for solving the inverse statics of a cable-driven soft arm with nonconstant curvature. \textit{IEEE Transactions on Robotics},  \textit{31}(4), 823-834. https://doi.org/10.1109/TRO.2015.2428511.

\bibitem{ML1}
Čakurda, T., Trojanová, M., Pomin, P., \& Hošovský, A. (2025). Deep learning methods in soft robotics: Architectures and applications. Advanced Intelligent Systems, 7(5), 2400576.

\bibitem{ML2}
Calzada-Garcia, A., Victores, J. G., Naranjo-Campos, F. J., \& Balaguer, C. (2025). A review on inverse kinematics, control and planning for robotic manipulators with and without obstacles via deep neural networks. Algorithms, 18(1), 23.

\bibitem{tim1}Haghshenas-Jaryani, M. (2023). Timoshenko beam-based analytical formulation and numerical simulation of continuum soft-bodied robotic arms. In \textit{2023 IEEE International Conference on Robotics and Biomimetics (ROBIO)}, 1-6, Koh Samui, Thailand, IEEE. https://doi.org/10.1109/ROBIO58561.2023.10354597

\bibitem{tim2}Fattahi, J. S., \& Spinello, D. (2013). Timoshenko beam model for exploration and sensing with a continuum centipede inspired robot. In \textit{Dynamic Systems and Control Conference}, Vol. 56123, V001T07A006. American Society of Mechanical Engineers. Palo Alto, USA.
https://doi.org/10.1115/DSCC2013-4103

\bibitem{tim3}Ling, M., Zhou, H., \& Chen, L. (2023). Dynamic stiffness matrix with Timoshenko beam theory and linear frequency solution for use in compliant mechanisms. \textit{Journal of Mechanisms and Robotics},  \textit{15}(6), 061002. https://doi.org/10.1115/1.4056236

\bibitem{tim4}Fattahi, J., \& Spinello, D. (2014). A Timoshenko beam reduced order model for shape tracking with a slender mechanism. \textit{Journal of Sound and Vibration},  \textit{333}(20), 5165-5180. https://doi.org/10.1016/j.jsv.2014.05.040

\bibitem{tim5}Wenlong, Y., Wei, D., \& Zhijiang, D. (2013). Mechanics-based kinematic modeling of a continuum manipulator. In \textit{2013 IEEE/RSJ International Conference on Intelligent Robots and Systems}, 5052-5058,  Tokyo, Japan, IEEE. https://doi.org/10.1109/IROS.2013.6697087

\bibitem{mod}
Singh, A., Pinto, M., Kaltsas, P., Pirozzi, S., Sulaiman, S., \& Ficuciello, F. (2024). Validations of various in-hand object manipulation strategies employing a novel tactile sensor developed for an under-actuated robot hand. Frontiers in Robotics and AI, 11, 1460589.

\bibitem{mod1}Sulaiman, S., \& Sudheer, A. P. (2021, April). Modelling of torso and dual arms for a humanoid robot with fixed base by using screw theory for dexterous applications. In IOP Conference Series: Materials Science and Engineering (Vol. 1132, No. 1, p. 012036). IOP Publishing.

\bibitem{mod2}Sulaiman, S., \& Sudheer, A. P. (2021). Dexterity analysis and intelligent trajectory planning of redundant dual arms for an upper body humanoid robot. Industrial Robot: the international journal of robotics research and application, 48(6), 915-928.

\bibitem{mod3}
Wang, T., Gao, Z., Li, C., Min, G., Xu, K., Zhao, E., ... \& Tang, W. (2026). Bioinspired textured sensor arrays with early temporal processing for ultrafast robotic tactile recognition. Materials Science and Engineering: R: Reports, 167, 101113.

\bibitem{mod4}
Mei, J., Ning, P., Ding, Y., Ni, J., \& Liu, P. (2026). A fuzzy-inference-based adaptive impedance control for Delta robots. Industrial Robot: the international journal of robotics research and application, 1-10.

\bibitem{mpc}
Schetter, F., Sulaiman, S., George, S., De Risi, P., \& Ficuciello, F. (2025, September). Optimizing Prosthetic Wrist Movement: A Model Predictive Control Approach. In International Conference on Social Robotics (pp. 239-252). Singapore: Springer Nature Singapore.

\bibitem{pros}Tinoco, V., Silva, M. F., Santos, F. N., Morais, R., Magalhães, S. A., \& Oliveira, P. M. (2025). A review of advanced controller methodologies for robotic manipulators. International Journal of Dynamics and Control,  \textit{13}(1), 36. https://doi.org/10.1007/s40435-024-01533-1

\bibitem{Skorina}Skorina, E. H., Luo, M., Ozel, S., Chen, F., Tao, W., \& Onal, C. D. (2015). Feedforward augmented sliding mode motion control of antagonistic soft pneumatic actuators. In \textit{2015 IEEE International Conference on Robotics and Automation (ICRA)}, 2544-2549, Seattle, USA, IEEE. https://doi.org/10.1109/ICRA.2015.7139540

\bibitem{khan}Khan, A. H., \& Li, S. (2020). Sliding mode control with PID sliding surface for active vibration damping of pneumatically actuated soft robots. \textit{IEEE Access}, \textit{8}, 88793-88800. https://doi.org/10.1109/ACCESS.2020.2992997

\bibitem{Mousa}Mousa, A., Khoo, S., \& Norton, M. (2018). Robust control of tendon driven continuum robots. In \textit{2018 15th International Workshop on Variable Structure Systems (VSS)}, 49-54, Graz, Austria, IEEE. https://doi.org/10.1109/VSS.2018.8460324

\bibitem{Al}Alqumsan, A. A., Khoo, S., Arogbonlo, A., \& Nahavand, S. (2021). Adaptive neural network based sliding mode control of continuum robots with mismatched uncertainties. In \textit{2021 IEEE International Conference on Systems, Man, and Cybernetics (SMC)}, 2602-2607, Melbourne, Australia, IEEE. https://doi.org/10.1109/SMC52423.2021.9659075

\bibitem{conf}
Sulaiman, S., Menon, M., Schetter, F., \& Ficuciello, F. (2024). Design, modelling, and experimental validation of a soft continuum wrist section developed for a prosthetic hand. In \textit{2024 IEEE/RSJ International Conference on Intelligent Robots and Systems (IROS)}, 11347-11354, Abu Dhabi, UAE, IEEE. https://doi.org/10.1109/IROS58592.2024.10802467

\bibitem{ref2}
Liu, H., Ferrentino, P., Pirozzi, S., Siciliano, B., \& Ficuciello, F. (2019). The PRISMA hand II: a sensorized robust hand for adaptive grasp and in-hand manipulation. In \textit{The International Symposium of Robotics Research}, 971-986, Hanoi, Vietnam, Cham: Springer International Publishing. https://doi.org/10.1007/978-3-030-95459-860

\bibitem{quasi}Tutcu, C., Baydere, B. A., Talas, S. K., \& Samur, E. (2021). Quasi-static modeling of a novel growing soft-continuum robot. \textit{The International Journal of Robotics Research}, \textit{40}(1), 86-98. https://doi.org/10.1177/0278364919893438

\bibitem{dynamic}Thomson, W. (2018). \textit{Theory of vibration with applications}. CrC Press.https://doi.org/10.1201/9780203718841










%\bibitem{fanny1}F. Ficuciello, "Synergy-Based Control of Underactuated Anthropomorphic Hands," in IEEE Transactions on Industrial Informatics, vol. 15, no. 2, pp. 1144-1152, Feb. 2019, doi: 10.1109/TII.2018.2841043.


\begin{comment}
\bibitem{de}de Souza, J. O. D. O., Bloedow, M. D., Rubo, F. C., de Figueiredo, R. M., Pessin, G., \& Rigo, S. J. (2021). Investigation of different approaches to real-time control of prosthetic hands with electromyography signals. \textit{IEEE  Sensors Journal}, 21(18), 20674-20684. https://doi.org/10.1109/JSEN.2021.3099744

\bibitem{palli}Palli, G., Melchiorri, C., Vassura, G., Scarcia, U., Moriello, L., Berselli, G., ... \& Siciliano, B. (2014). The DEXMART hand: Mechatronic design and experimental evaluation of synergy-based control for human-like grasping. \textit{The International Journal of Robotics Research}, 33(5), 799-824. https://doi.org/10.1177/0278364913519897.

\bibitem{ref1}
Chen, Z., Min, H., Wang, D., Xia, Z., Sun, F., \& Fang, B. (2023). A review of myoelectric control for prosthetic hand manipulation. \textit{Biomimetics}, 8(3), 328.  https://doi.org/10.3390/biomimetics8030328

\bibitem{ref_2}
Gentile, C., Cordella, F., \& Zollo, L. (2022). Hierarchical human-inspired control strategies for prosthetic hands. \textit{Sensors}, 22(7), 2521. https://doi.org/10.3390/s22072521



\bibitem{ref_3}
Marasco, P. D., Hebert, J. S., Sensinger, J. W., Shell, C. E., Schofield, J. S., Thumser, Z. C., ... \& Orzell, B. M. (2018). Illusory movement perception improves motor control for prosthetic hands. \textit{Science translational medicine}, 10(432), eaao6990. https://doi.org/10.1126/scitranslmed.aao6990

%\bibitem{review}Armanini C, Boyer F, Mathew AT, Duriez C, Renda F. Soft robots modeling: A structured overview. IEEE Transactions on Robotics. 2023 Jan 6;39(3):1728-48.



\bibitem{w1}
Legrand, M., Jarrassé, N., Richer, F., \& Morel, G. (2020, May). A closed-loop and ergonomic control for prosthetic wrist rotation. In \textit{2020 IEEE International Conference on Robotics and Automation (ICRA)}, 2763-2769. IEEE. https://doi.org/10.1109/ICRA40945.2020.9197554

\bibitem{w2}
Swami, C. P., Lenhard, N., \& Kang, J. (2021). A novel framework for designing a multi-DoF prosthetic wrist control using machine learning. \textit{Scientific Reports}, 11(1), 15050.  https://doi.org/10.1038/s41598-021-94449-1

\bibitem{w3}
Ma, J., Thakor, N. V., \& Matsuno, F. (2014). Hand and wrist movement control of myoelectric prosthesis based on synergy. \textit{IEEE Transactions on Human-Machine Systems}, 45(1), 74-83. https://doi.org/10.1109/THMS.2014.2358634

\bibitem{w4}
Bennett, D. A., \& Goldfarb, M. (2017). IMU-based wrist rotation control of a transradial myoelectric prosthesis. \textit{IEEE Transactions on Neural Systems and Rehabilitation Engineering}, 26(2), 419-427. https://doi.org/10.1109/TNSRE.2017.2682642

\bibitem{w5}
Epp, J. G., Ours, M. K., Chaplinski, P. A., \& Pesch, A. H. (2024). Design and Control of a Prosthetic Wrist Mechanism. In \textit{International Design Engineering Technical Conferences and Computers and Information in Engineering Conference}, Vol. 88414,  V007T07A029. American Society of Mechanical Engineers. 
https://doi.org/10.1115/DETC2024-143057

\bibitem{w6}
Engeberg, E., Frankel, M., \& Meek, S. (2009). Biomimetic grip force compensation based on acceleration of a prosthetic wrist under sliding mode control. In \textit{2008 IEEE International Conference on Robotics and Biomimetics},  210-215. IEEE. https://doi.org/10.1109/ROBIO.2009.4913005

\bibitem{w7}
Abbasi Moshaii, A., Mohammadi Moghaddam, M., \& Dehghan Niestanak, V. (2019). Fuzzy sliding mode control of a wearable rehabilitation robot for wrist and finger. \textit{Industrial Robot: the international journal of robotics research and application}, 46(6), 839-850. https://doi.org/10.1108/IR-05-2019-0110

\bibitem{cao}Cao, G., Liu, Y., Jiang, Y., Zhang, F., Bian, G., \& Owens, D. H. (2021). Observer-based continuous adaptive sliding mode control for soft actuators. \textit{Nonlinear Dynamics}, 105(1), 371-386. https://doi.org/10.1007/s11071-021-06606-w

\bibitem{Kazemipour}Kazemipour, A., Fischer, O., Toshimitsu, Y., Wong, K. W., \& Katzschmann, R. K. (2022). Adaptive dynamic sliding mode control of soft continuum manipulators. In \textit{2022 International Conference on Robotics and Automation (ICRA)}, 3259-3265. IEEE. https://doi.org/10.1109/ICRA46639.2022.9811715

\bibitem{ref_4}
Engeberg, E. D., \& Meek, S. G. (2011). Adaptive sliding mode control for prosthetic hands to simultaneously prevent slip and minimize deformation of grasped objects. \textit{IEEE/ASME Transactions on Mechatronics}, 18(1), 376-385. https://doi.org/10.1109/TMECH.2011.2179061

\bibitem{ref_5}
Engeberg, E. D., \& Meek, S. G. (2011). Adaptive sliding mode control of grasped object slip for prosthetic hands. In \textit{2011 IEEE/RSJ International Conference on Intelligent Robots and Systems}, 4174-4179. IEEE. https://doi.org/10.1109/IROS.2011.6094500

\bibitem{ref_6}
Engeberg, E. D. (2010). Biomimetic sliding mode control of a prosthetic hand. In \textit{2010 3rd IEEE RAS \& EMBS International Conference on Biomedical Robotics and Biomechatronics}, 343-348. IEEE. https://doi.org/10.1109/BIOROB.2010.5626815

\bibitem{ref_7}
Majeed, H. S., Kadhim, S. K., \& Jaber, A. A. (2022). Design of a sliding mode controller for a prosthetic human hand’s finger. \textit{Engineering and Technology Journal}, 40(1), 257-266. 

\bibitem{ref_8}
Islam, M. R., Rahmani, M., \& Rahman, M. H. (2020). A novel exoskeleton with fractional sliding mode control for upper limb rehabilitation. \textit{Robotica}, 38(11), 2099-2120. https://doi.org/10.1017/S0263574719001851 


\end{comment}

\end{thebibliography}
\end{document}